# Semi-Supervised Learning of Class Balance
# under Class-Prior Change by Distribution Matching

**Marthinus Christoffel du Plessis**  CHRISTO@SG.CS.TITECH.AC.JP
**Masashi Sugiyama**  SUGI@SG.CS.TITECH.AC.JP
Department of Computer Science, Tokyo Institute of Technology, Tokyo, Japan

## Abstract

In real-world classification problems, the class balance in the training dataset does not necessarily reflect that of the test dataset, which can cause significant estimation bias. If the class ratio of the test dataset is known, instance re-weighting or resampling allows systematical bias correction. However, learning the class ratio of the test dataset is challenging when no labeled data is available from the test domain. In this paper, we propose to estimate the class ratio in the test dataset by matching probability distributions of training and test input data. We demonstrate the utility of the proposed approach through experiments.

## 1. Introduction

Most supervised learning algorithms assume that training and test data follow the same probability distribution (Vapnik, 1998; Hastie et al., 2001; Bishop, 2006). However, this de facto standard assumption is often violated in real-world problems, caused by intrinsic sample selection bias or inevitable non-stationarity (Heckman, 1979; Quiñonero-Candela et al., 2009; Sugiyama & Kawanabe, 2012).

In classification scenarios, changes in class balance are often observed—for example, the male-female ratio is almost fifty-fifty in the real-world (test set), whereas training samples collected in a research laboratory tends to be dominated by male data. Such a situation is called a *class-prior change*, and the bias caused by differing class balances can be systematically adjusted by instance re-weighting or resampling if the class balance in the test dataset is known (Elkan, 2001; Lin et al., 2002).

However, the class ratio in the test dataset is often unknown in practice. A possible approach to coping with this problem is to learn a classifier so that the performance for all possible class balances is improved, e.g., through maximization of the area under the ROC curve (Cortes & Mohri, 2004; Clémençon et al., 2009). Another, possibly more direct approach is to estimate the class ratio in the test dataset and use the estimates for instance re-weighting or resampling. In this paper, we focus on the latter scenario under a semi-supervised learning setup (Chapelle et al., 2006), where no labeled data is available from the test domain.

Saerens et al. (2001) is a seminal paper on this topic, which proposed to estimate the class ratio by the expectation-maximization (EM) algorithm (Dempster et al., 1977)—alternately updating the test class-prior and class-posterior probabilities from some initial estimates until convergence. This method has been successfully applied to various real-world problems such as word sense disambiguation (Chan & Ng, 2006) and remote sensing (Latinne et al., 2001).

In this paper, we first reformulate the above algorithm, and show that this actually corresponds to approximating the test input distribution by a linear combination of class-wise input distributions under the Kullback-Leibler (KL) divergence (Kullback & Leibler, 1951). In this procedure, the class-wise input distributions are approximated via class-posterior estimation, for example, by kernel logistic regression (Hastie et al., 2001) or its squared-loss variant (Sugiyama, 2010).

This new formulation motivates us to develop a new approach, since indirectly estimating the divergence by estimating the individual class-posterior distributions may not be the best scheme. Recently, KL divergence estimation based on *direct density-ratio estimation* has been shown to be promising (Nguyen et al., 2010; Sugiyama et al., 2008). Furthermore, a squared-loss variant of the KL divergence called the Pearson (PE) divergence (Pearson, 1900) can also be approximated in the same way, with an analytic solution that can be computed efficiently (Kanamori et al., 2009a). The PE divergence and the KL divergence both belong to the *f*-divergence class (Ali & Silvey, 1966; Csiszár, 1967), which share similar properties. In this paper, with the aid of this density-ratio based PE divergence estimator, we propose a new semi-supervised method for estimating the class ratio in the test dataset. Through experiments, we demonstrate the usefulness of the proposed method.

## 2. Problem Formulation and Existing Method

In this section, we formulate the problem of semi-supervised class-prior estimation and review an existing method (Saerens et al., 2001).

### 2.1. Problem Formulation

Let $\boldsymbol{x} \in \mathbb{R}^d$ be the $d$-dimensional input data, $y \in \{1, \ldots, c\}$ be the class label, and $c$ be the number of classes. We consider class-prior change, i.e., the class-prior probability for training data $p(y)$ and that for test data $p'(y)$ are different. However, we assume that the class-conditional density for training data $p(\boldsymbol{x}|y)$ and that for test data $p'(\boldsymbol{x}|y)$ are the same:

$$p(\boldsymbol{x}|y) = p'(\boldsymbol{x}|y). \tag{1}$$

Note that training and test joint densities $p(\boldsymbol{x}, y)$ and $p'(\boldsymbol{x}, y)$ as well as training and test input densities $p(\boldsymbol{x})$ and $p'(\boldsymbol{x})$ are generally different under this setup.

The goal of this paper is to estimate $p'(y)$ from labeled training samples $\{(\boldsymbol{x}_i, y_i)\}_{i=1}^n$ drawn independently from $p(\boldsymbol{x}, y)$ and unlabeled test samples $\{\boldsymbol{x}_i'\}_{i=1}^{n'}$ drawn independently from $p'(\boldsymbol{x})$. Given test labels $\{y_i'\}_{i=1}^{n'}$, $p'(y)$ can be naively estimated by $n_y'/n'$, where $n_y'$ is the number of test samples in class $y$. Here, however, we would like to estimate $p'(y)$ *without* $\{y_i'\}_{i=1}^{n'}$.

### 2.2. Existing Method

We give a brief overview of an existing method for semi-supervised class-prior estimation (Saerens et al., 2001), which is based on the expectation-maximization (EM) algorithm (Dempster et al., 1977).

In the algorithm, test class-prior and class-posterior estimates $\widehat{p}'(y)$ and $\widehat{p}'(y|\boldsymbol{x})$ are iteratively updated as follows:

1. Obtain an estimate of the training class-posterior probability, $\widehat{p}(y|\boldsymbol{x})$, from training data $\{(\boldsymbol{x}_i, y_i)\}_{i=1}^n$, for example, by kernel logistic regression (Hastie et al., 2001) or its squared-loss variant (Sugiyama, 2010).

2. Obtain an estimate of the training class-prior probability, $\widehat{p}(y)$, from the labeled training data $\{(\boldsymbol{x}_i, y_i)\}_{i=1}^n$ as $\widehat{p}(y) = n_y/n$, where $n_y$ is the number of training samples in class $y$. Set the initial estimate of the test class-posterior probability equal to it: $\widehat{p}_0'(y) = \widehat{p}(y)$.

3. Repeat until convergence: $t = 1, 2, \ldots$

   (a) Compute a new test class-posterior estimate $\widehat{p}_t'(y|\boldsymbol{x})$ based on the current test class-prior estimate $\widehat{p}_{t-1}'(y)$ as

   $$\widehat{p}_t'(y|\boldsymbol{x}) = \frac{\widehat{p}_{t-1}'(y)\widehat{p}(y|\boldsymbol{x})/\widehat{p}(y)}{\sum_{y'=1}^c \widehat{p}_{t-1}'(y')\widehat{p}(y'|\boldsymbol{x})/\widehat{p}(y')}. \tag{2}$$

   (b) Compute a new test class-prior estimate $\widehat{p}_t'(y)$ based on the current test class-prior estimate $\widehat{p}_t'(y|\boldsymbol{x})$ as

   $$\widehat{p}_t'(y) = \frac{1}{n'}\sum_{i=1}^{n'} \widehat{p}_t'(y|\boldsymbol{x}_i'). \tag{3}$$

This procedure was shown to converge to a local optimal solution.

Note that Eq.(2) comes from the Bayes formulae,

$$p(\boldsymbol{x}|y) = \frac{p(y|\boldsymbol{x})p(\boldsymbol{x})}{p(y)} \text{ and } p'(\boldsymbol{x}|y) = \frac{p'(y|\boldsymbol{x})p'(\boldsymbol{x})}{p'(y)},$$

combined with Eq.(1):

$$p'(y|\boldsymbol{x}) \propto \frac{p'(y)}{p(y)}p(y|\boldsymbol{x}).$$

Eq.(3) comes from empirical marginalization of

$$p'(y) = \int p'(y|\boldsymbol{x})p'(\boldsymbol{x})\mathrm{d}\boldsymbol{x}.$$

## 3. Reformulation of the EM Algorithm as Distribution Matching

In this section, we show that the above EM algorithm can be interpreted as matching the test input density to a linear combination of class-wise input distributions under the Kullback-Leibler (KL) divergence (Kullback & Leibler, 1951).

Based on the assumption that the class-conditional densities for training and test data are unchanged (see Eq.(1)), let us model the test input density $p'(\boldsymbol{x})$ by

$$q'(\boldsymbol{x}) = \sum_{y=1}^c \theta_y p(\boldsymbol{x}|y), \tag{4}$$

where $\theta_y$ is a coefficient corresponding to $p'(y)$:

$$\sum_{y=1}^c \theta_y = 1. \tag{5}$$

We match the model $q'(\boldsymbol{x})$ with the test input density $p'(\boldsymbol{x})$ under the KL divergence:

$$
\begin{aligned}
\mathrm{KL}(p'\|q') &:= \int p'(\boldsymbol{x}) \log \frac{p'(\boldsymbol{x})}{q'(\boldsymbol{x})} \mathrm{d}\boldsymbol{x} \\
&= \int p'(\boldsymbol{x}) \log p'(\boldsymbol{x}) \mathrm{d}\boldsymbol{x} \\
&\quad - \int p'(\boldsymbol{x}) \log \left( \sum_{y=1}^{c} \theta_y p(\boldsymbol{x}|y) \right) \mathrm{d}\boldsymbol{x}. \quad (6)
\end{aligned}
$$

Ignoring the first term (which is a constant) and approximating the expectation in the second term with its empirical average give the following optimization problem:

$$
\max_{\{\theta_y\}_{y=1}^{c}} \frac{1}{n'} \sum_{i=1}^{n'} \log \left( \sum_{y=1}^{c} \theta_y p(\boldsymbol{x}_i'|y) \right), \quad (7)
$$

subject to Eq.(5).

Since the above maximization is a convex optimization problem, the Karush-Kuhn-Tucker (KKT) conditions are necessary and sufficient for optimality (Boyd & Vandenberghe, 2004). The KKT conditions for the above problem is given by Eq.(5) and

$$
\frac{1}{n'} \sum_{i=1}^{n'} \frac{p(\boldsymbol{x}_i'|y)}{\sum_{y'=1}^{c} \theta_{y'} p(\boldsymbol{x}_i'|y')} = \nu, \quad \forall y = 1, \dots, c,
$$

where $\nu$ is a Lagrange multiplier. From these equations, we can determine $\nu$ as

$$
\begin{aligned}
\nu &= 1 \cdot \nu = \left( \sum_{y=1}^{c} \theta_y \right) \cdot \left( \frac{1}{n'} \sum_{i=1}^{n'} \frac{p(\boldsymbol{x}_i'|y)}{\sum_{y'=1}^{c} \theta_{y'} p(\boldsymbol{x}_i'|y')} \right) \\
&= \frac{1}{n'} \sum_{i=1}^{n'} \frac{\sum_{y=1}^{c} \theta_y p(\boldsymbol{x}_i'|y)}{\sum_{y'=1}^{c} \theta_{y'} p(\boldsymbol{x}_i'|y')} = 1.
\end{aligned}
$$

Then the solution $\{\theta_y\}_{y=1}^{c}$ can be calculated by fixed-point iteration as follows (McLachlan & Krishnan, 1997):

$$
\theta_y \leftarrow \theta_y \left( \frac{1}{n'} \sum_{i=1}^{n'} \frac{p(\boldsymbol{x}_i'|y)}{\sum_{y=1}^{c} \theta_y p(\boldsymbol{x}_i'|y)} \right). \quad (8)
$$

Making the substitution $p(\boldsymbol{x}_i'|y) = p(y|\boldsymbol{x}_i')p(\boldsymbol{x}_i')/p(y)$, canceling $p(\boldsymbol{x}_i')$ in the numerator and denominator, and replacing $p(y|\boldsymbol{x})$ with $\widehat{p}(y|\boldsymbol{x})$, we can show that the above updating formula is reduced to

$$
\theta_y \leftarrow \frac{1}{n'} \sum_{i=1}^{n'} \frac{\theta_y \widehat{p}(y|\boldsymbol{x}_i')/\widehat{p}(y)}{\sum_{y'=1}^{c} \theta_{y'} \widehat{p}(y'|\boldsymbol{x}_i')/\widehat{p}(y')},
$$

which is the same as Eq.(3) with Eq.(2) substituted.

Therefore, the EM method is essentially equivalent to matching the training and test input distributions under the KL divergence, which uses the class-conditional density $p(\boldsymbol{x}|y)$ as a building block (see Eq.(8)). However, this fact is not apparent in the EM expression because of the cancellation of $p(\boldsymbol{x}_i')$ in the numerator and denominator.

The convexity of Eq.(7) implies that there are no local minima. However, this was not recognized in Saerens et al. (2001) since the algorithm was derived via the incomplete data EM method.

## 4. Class-Prior Estimation by Direct Divergence Minimization

The analysis in the previous section motivates us to explore a more direct way to learn coefficients $\{\theta_y\}_{y=1}^{c}$. That is, given an estimator of a divergence from $p'$ to $q'$, coefficients $\{\theta_y\}_{y=1}^{c}$ are learned so that the divergence estimator is minimized.

In this section, we first review a general framework of approximating the *f-divergences* (Ali & Silvey, 1966; Csiszár, 1967) via *Legendre-Fenchel convex duality* (Keziou, 2003; Nguyen et al., 2010). Then we review two specific methods of divergence estimation for the KL divergence and the Pearson (PE) divergence (Pearson, 1900). Finally, we propose to use the PE divergence estimator for determining the coefficients $\{\theta_y\}_{y=1}^{c}$.

### 4.1. Framework of $f$-Divergence Approximation

An $f$-divergence (Ali & Silvey, 1966; Csiszár, 1967) from $p'$ to $q'$ is a general divergence measure defined by a convex function $f$ such that $f(1) = 0$ as

$$
D_f(p'\|q') := \int p'(\boldsymbol{x}) f\left( \frac{q'(\boldsymbol{x})}{p'(\boldsymbol{x})} \right) \mathrm{d}\boldsymbol{x}.
$$

It was shown that the $f$-divergence can be lower-bounded via *Legendre-Fenchel convex duality* (Rockafellar, 1970) as follows (Keziou, 2003; Nguyen et al., 2010):

$$
\begin{aligned}
D_f(p'\|q') = \max_{r} \Bigg[ &\int q'(\boldsymbol{x}) r(\boldsymbol{x}) \mathrm{d}\boldsymbol{x} \\
&- \int p'(\boldsymbol{x}) f^*(r(\boldsymbol{x})) \mathrm{d}\boldsymbol{x} \Bigg], \quad (9)
\end{aligned}
$$

where $f^*$ is the convex conjugate of $f$. The maximum is achieved if and only if $r(\boldsymbol{x}) = q'(\boldsymbol{x})/p'(\boldsymbol{x})$. Eq.(9) is a useful expression because the right-hand side only contains expectations of $r$ and $f^*(r(\boldsymbol{x}))$, which can be simply approximated by sample averages.

Below, we show specific methods of divergence approximation for the KL and PE divergences under model (4)

and the following parametric expression of the density ratio $r(\boldsymbol{x})$:

$$r(\boldsymbol{x}) = \sum_{\ell=0}^{b} \alpha_\ell \varphi_\ell(\boldsymbol{x}), \qquad (10)$$

where $\{\alpha_\ell\}_{\ell=0}^{b}$ are parameters and $\{\varphi_\ell(\boldsymbol{x})\}_{\ell=0}^{b}$ are basis functions. In practice, we use a constant basis and Gaussian kernels centered at the training data points, i.e., for $b = n$ and $\ell = 1, 2, \ldots, n$,

$$\varphi_0(\boldsymbol{x}) = 1 \quad \text{and} \quad \varphi_\ell(\boldsymbol{x}) = \exp\left(-\frac{\|\boldsymbol{x} - \boldsymbol{x}_\ell\|^2}{2\sigma^2}\right).$$

This provides a non-parametric divergence estimator (Nguyen et al., 2010; Sugiyama et al., 2008; Kanamori et al., 2012).

### 4.2. KL-Divergence Approximation

With $f(u) = -\log u$ for $u > 0$ and $+\infty$ for $u \le 0$, the $f$-divergence is reduced to the KL divergence. For this $f$, the convex conjugate is given by $f^*(v) = -1 - \log(-v)$ for $v < 0$ and $+\infty$ for $v \ge 0$. Then, if $-\alpha_\ell$ is regarded as $\alpha_\ell$, an empirical approximation of Eq.(9) under (4) and (10) is given as follows (Nguyen et al., 2010):

$$\begin{aligned}
\text{KL}(p' \| q') \approx \max_{\{\alpha_\ell\}_{\ell=0}^{b}} \Bigg[ &-\sum_{y=1}^{c} \frac{\theta_y}{n_y} \sum_{i:y_i=y} \sum_{\ell=0}^{b} \alpha_\ell \varphi_\ell(\boldsymbol{x}_i) \\
&+ \frac{1}{n'} \sum_{i=1}^{n'} \log\left(\sum_{\ell=0}^{b} \alpha_\ell \varphi_\ell(\boldsymbol{x}_i')\right) + 1 \Bigg],
\end{aligned}$$

subject to $\alpha_0, \alpha_1, \ldots, \alpha_b \ge 0$. A similar approach, which directly estimates the inverted ratio $p'(\boldsymbol{x})/q'(\boldsymbol{x})$ with the same model (10), is also known (Sugiyama et al., 2008):

$$\text{KL}(p' \| q') \approx \max_{\{\alpha_\ell\}_{\ell=0}^{b}} \left[ \frac{1}{n'} \sum_{i=1}^{n'} \log\left(\sum_{\ell=0}^{b} \alpha_\ell \varphi_\ell(\boldsymbol{x}_i')\right) \right],$$

subject to $\alpha_0, \alpha_1, \ldots, \alpha_b \ge 0$ and

$$\sum_{y=1}^{c} \frac{\theta_y}{n_y} \sum_{i:y_i=y} \sum_{\ell=0}^{b} \alpha_\ell \varphi_\ell(\boldsymbol{x}_i) = 1.$$

These are convex optimization problems, and thus global optimal solutions can be obtained by naive optimization. Tuning parameters possibly included in the basis function such as the kernel width can be systematically optimized by cross-validation (Sugiyama et al., 2008). The KL-divergence estimator obtained above was proved to possess superior convergence properties both in parametric and non-parametric setups (Sugiyama et al., 2008; Nguyen et al., 2010).

However, computing the KL-divergence estimator is rather time-consuming because optimization of $\{\alpha_\ell\}_{\ell=0}^{b}$ needs to be carried out for each $\{\theta_y\}_{y=1}^{c}$.

### 4.3. PE-Divergence Approximation

As an alternative to the KL-divergence, let us consider the PE divergence defined by

$$\text{PE}(p' \| q') := \frac{1}{2} \int \left(\frac{q'(\boldsymbol{x})}{p'(\boldsymbol{x})} - 1\right)^2 p'(\boldsymbol{x}) \mathrm{d}\boldsymbol{x}, \qquad (11)$$

which is a squared-loss variant of the KL divergence and is a $f$-divergence with $f(u) = (t-1)^2/2$.

For this $f$, the convex conjugate is given by $f^*(v) = v^2/2 + v$. Then, an empirical approximation of Eq.(9) under (4) and (10) is given as follows (Kanamori et al., 2009a):

$$\text{PE}(p' \| q') \approx \max_{\boldsymbol{\alpha}} \left[ -\frac{1}{2} \boldsymbol{\alpha}^\top \widehat{\boldsymbol{G}} \boldsymbol{\alpha} + \boldsymbol{\alpha}^\top \widehat{\boldsymbol{H}} \boldsymbol{\theta} - \frac{1}{2} \right],$$

where

$$\boldsymbol{\alpha} = [\alpha_0 \ \alpha_1 \ \cdots \ \alpha_b]^\top, \quad \widehat{\boldsymbol{G}} = \frac{1}{n'} \sum_{i=1}^{n'} \boldsymbol{\varphi}(\boldsymbol{x}_i') \boldsymbol{\varphi}(\boldsymbol{x}_i')^\top,$$

$$\boldsymbol{\varphi}(\boldsymbol{x}) = [\varphi_0(\boldsymbol{x}) \ \varphi_1(\boldsymbol{x}) \ \cdots \ \varphi_b(\boldsymbol{x})], \quad \widehat{\boldsymbol{H}} = \left[\widehat{\boldsymbol{h}}_1 \ \cdots \ \widehat{\boldsymbol{h}}_c\right],$$

$$\widehat{\boldsymbol{h}}_y = \frac{1}{n_y} \sum_{i:y_i=y} \boldsymbol{\varphi}(\boldsymbol{x}_i), \quad \boldsymbol{\theta} = [\theta_1 \ \theta_2 \ \cdots \ \theta_c]^\top.$$

A regularized solution to the above maximization problem can be obtained analytically as

$$\widehat{\boldsymbol{\alpha}} = \left(\widehat{\boldsymbol{G}} + \lambda \boldsymbol{R}\right)^{-1} \widehat{\boldsymbol{H}} \boldsymbol{\theta}, \qquad (12)$$

where $\lambda$ is a positive constant and $\boldsymbol{R}$ is defined as

$$\boldsymbol{R} = \begin{bmatrix} 0 & \boldsymbol{0}_{1 \times b} \\ \boldsymbol{0}_{b \times 1} & \boldsymbol{I}_{b \times b} \end{bmatrix}.$$

The PE divergence estimator obtained above was proved to have superior convergence properties both in parametric and non-parametric setups (Kanamori et al., 2009a; 2012). Tuning parameters possibly included in the basis function such as the kernel width or the regularization parameter can be systematically optimized by cross-validation (Kanamori et al., 2009a; 2012).

### 4.4. Learning Class Ratios by PE Divergence Matching

As shown above, the KL and PE divergences can be systematically estimated without density estimation via Legendre-Fenchel convex duality. Among them, the PE divergence estimator, explicitly expressed as

$$\begin{aligned}
\widehat{\text{PE}}(\boldsymbol{\theta}) := &-\frac{1}{2} \boldsymbol{\theta}^\top \widehat{\boldsymbol{H}}^\top \left(\widehat{\boldsymbol{G}} + \lambda \boldsymbol{R}\right)^{-1} \widehat{\boldsymbol{G}} \left(\widehat{\boldsymbol{G}} + \lambda \boldsymbol{R}\right)^{-1} \widehat{\boldsymbol{H}} \boldsymbol{\theta} \\
&+ \boldsymbol{\theta}^\top \widehat{\boldsymbol{H}}^\top \left(\widehat{\boldsymbol{G}} + \lambda \boldsymbol{R}\right)^{-1} \widehat{\boldsymbol{H}} \boldsymbol{\theta} - \frac{1}{2},
\end{aligned}$$

is more useful for our purpose of learning class ratios, because of the following reasons: The PE-divergence was shown to be more robust against outliers than the KL-divergence, based on power divergence analysis (Basu et al., 1998; Sugiyama et al., 2012). This is a useful property in practical data analysis suffering high noise and outliers. Furthermore, the above PE-divergence estimator was shown to possess the minimum condition number among a general class of estimators, meaning that it is the most stable estimator (Kanamori et al., 2009b).

Another, and practically more important advantage of the above PE divergence estimator is that it can be computed efficiently and analytically. This advantage is even more crucial in our case because we minimize the above PE divergence estimator with respect to $\boldsymbol{\theta}$:

$$\min_{\boldsymbol{\theta}} \widehat{\mathrm{PE}}(\boldsymbol{\theta})$$

$$\text{subject to } \sum_{y=1}^{c} \theta_y = 1 \text{ and } \theta_1, \ldots, \theta_c \geq 0.$$

Because $\widehat{\mathrm{PE}}(\boldsymbol{\theta})$ is given analytically as a function of $\boldsymbol{\theta}$, we can easily obtain the minimizer $\widehat{\boldsymbol{\theta}}$ by simple optimization strategies such as alternate gradient descent and projection or just a grid search, without re-computing the PE divergence estimator.

# 5. Experiments

In this section, we report experimental results.

## 5.1. Setup

The following five methods are compared:

- **EM-KLR**: The method of Saerens et al. (2001) (see Section 2.2). The class-posterior probability of the training dataset is estimated using $\ell_2$-penalized kernel logistic regression with Gaussian kernels. The L-BFGS quasi-Newton implementation included in the 'minFunc' package is used for logistic regression training (Schmidt, 2005).

- **KL-KDE**: The KL divergence estimator based on kernel density estimation (KDE). The class-wise input densities are estimated by KDE with Gaussian kernels. The kernel widths are estimated using likelihood cross-validation (Silverman, 1986).

- **PE-KDE**: The PE divergence estimator based on KDE. The class-wise input densities are estimated by KDE with Gaussian kernels. The kernel widths are estimated using least-squares cross-validation (Silverman, 1986).

Table 1. Datasets used in the experiments.

| Dataset | $d$ | # samples | # positives | # negatives |
|---|---|---|---|---|
| Australian | 14 | 690 | 307 | 383 |
| Diabetes | 8 | 768 | 500 | 268 |
| German | 24 | 1000 | 300 | 700 |
| Ionosphere | 34 | 351 | 225 | 126 |
| SAHeart | 9 | 462 | 302 | 160 |
| Twonorm | 20 | 7400 | 3697 | 3703 |

- **KL-DR**: The proposed method (see Section 4.2) using a KL divergence estimator based on the density ratio (DR). For the optimization, the L-BFGS with projection implementation 'minFuncBC' is used (Schmidt, 2005).

- **PE-DR**: The proposed method (see Section 4.4) using the PE divergence estimator based on DR.

Below, we compare accuracy of class-prior estimation and classification.

## 5.2. Benchmark Datasets

Here, we use binary-classification benchmark datasets listed in Table 1. We select 10 samples from each of the two classes for the training dataset and 50 samples for the test dataset. The samples in the test set are selected with probability $\theta^*$ from the first class and $(1 - \theta^*)$ from the second class, where $\theta^* = 0.1, 0.2, 0.3, 0.4, 0.5$.

The average squared error of the estimated class ratios are given in Figure 1. This shows that methods based on the KL and PE divergences overall outperform EM-KLR, implying that our reformulation of the EM algorithm as distribution matching (see Section 3) contributes to obtaining accurate class-ratio estimates. Among the KL-based methods, KL-KDE tends to perform better than KL-DR. This is because, in KL-KDE, we did not estimate the first term in Eq.(6), which is the negative entropy and is a constant. On the other hand, the negative entropy is also implicitly estimated in KL-DR, possibly incurring additional estimation error. Among the PE-based methods, PE-DR outperforms PE-KDE, showing that directly estimating density ratios without density estimation is more promising as a PE divergence estimator. Overall, PE-DR is shown to be the most accurate.

Next, we compare classification accuracy when the learned class-prior probabilities are used as instance weights. Figure 2 shows misclassification rates for a regularized least-squares classifier (Rifkin et al., 2003) with instance weighting. The results show that, as expected, a more accurate estimate of the class ratio tends to give a lower misclassification rate.

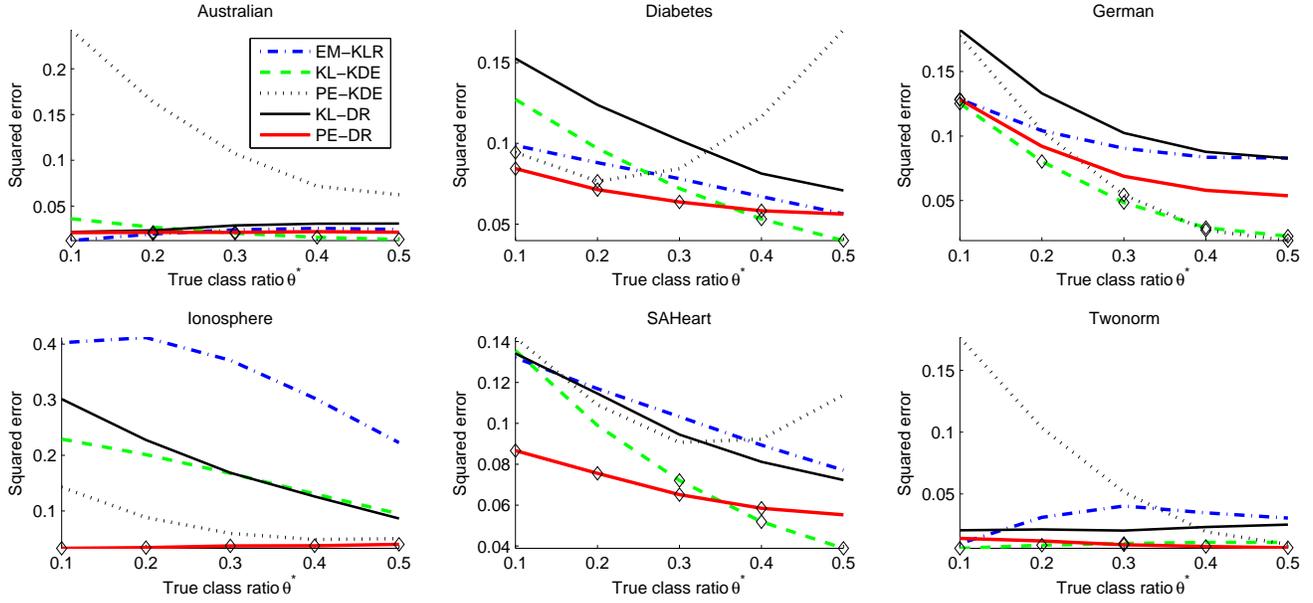

Figure 1. Average squared error between the true class ratio $\theta^*$ and estimated class ratio $\widehat{\theta}$ for the benchmark datasets listed in Table 1. The best method and comparable methods according to the t-test at significance level of 5% are indicated with a '⋄'

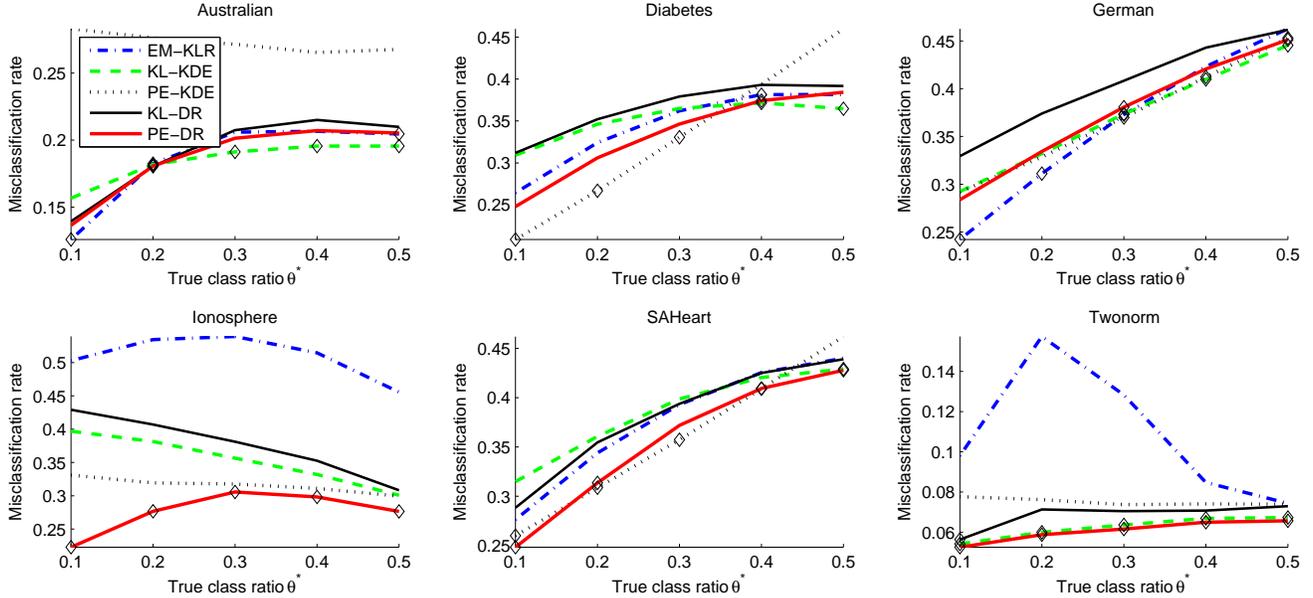

Figure 2. Average misclassification rates for the datasets listed in Table 1. Classification is performed using a regularized least-squares classifier with instance weighting. The best method and comparable methods according to the t-test at significance level of 5% are indicated with a '⋄'.

### 5.3. Real-World Application

Finally, we demonstrate the usefulness of the proposed approach in a real-world problem of military vehicle classification from geophone recordings (Duarte & Hu, 2004). This is a three class problem: Two vehicle classes and a class of recorded noise. The features are 50-dimensional. In this vehicle classification task, class-prior change is in-evitable because the type of vehicles passing through differs depending on time (e.g., day and night).

$n$ samples are drawn from each of the labeled classes for the training set with the uniform class prior, whereas 100 samples are drawn with probabilities $p = [0.6 \ 0.1 \ 0.3]$ from each of the classes for the test set. Due to the prohibitive computational cost, KL-DR was not included in this exper-iment.

In Figure 3, we plot the $\ell_2$-distance between the true and estimated class priors and the misclassification rate based on instance-weighted kernel logistic regression (Hastie et al., 2001) averaged over 1000 runs as functions of the number of training samples. As can be seen from the graphs, the performance of all methods improves as the number of training samples increases. Among the compared methods, PE-DR provides the most accurate estimates of the class prior and thus yields the lowest classification error.

## 6. Conclusion

Class-prior change is a problem that is conceivable in many real-world datasets, and it can be systematically corrected for if the class-prior of the test dataset is known. In this paper, we discussed the problem of estimating the test class ratios under the semi-supervised learning setup.

We first showed that the EM-based estimator introduced in Saerens et al. (2001) can be regarded as indirectly matching the test input distribution by a linear combination of class-wise input distributions. Based on this view, we proposed to use an explicit and possibly more accurate divergence estimator based on density-ratio estimation (Kanamori et al., 2009a) for learning test class-priors. The proposed method was shown to have various nice properties such as high robustness to noise and outliers, superior numerical stability, and excellent computational efficiency. Through experiments, we showed that the class ratios estimated by the proposed method are more accurate than competing methods, which can be translated into better classification accuracy.

## Acknowledgments

The authors thank the anonymous reviewers for their helpful comments. MCdP was supported by the MEXT scholarship, and MS was supported by AOARD and JST PRESTO.

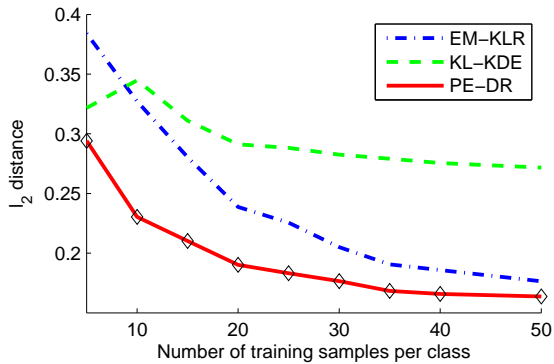

(a) The $\ell_2$-distance between the true and estimated class priors.

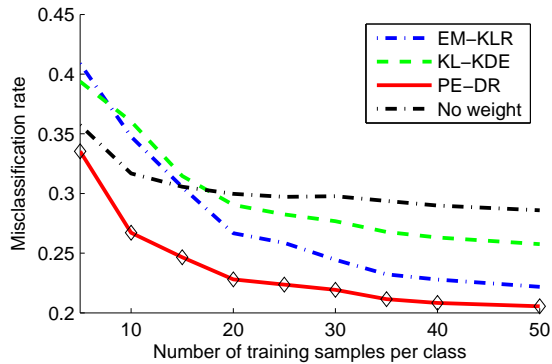

(b) Misclassification rate with instance-weighted kernel logistic regression.

*Figure 3.* Experimental results for the vehicle classification problem. The best method and comparable methods according to the t-test at significance level of 5% are indicated with a '⋄'.